\documentclass{article}

\usepackage{url}
\usepackage{graphicx}
\usepackage{multirow}
\usepackage{amsmath} 
\usepackage{amssymb}  
\usepackage{algorithmicx}
\usepackage{algpseudocode}
\usepackage{algorithm}

\usepackage[utf8]{inputenc}
\usepackage[T1]{fontenc}

\DeclareGraphicsExtensions{.pdf,.jpeg,,jpg,.png}
\graphicspath{{./figs/}}

\begin{document}
\title{Combining Fuzzy Cognitive Maps and Discrete Random Variables
}
%
%
\author{Piotr Szwed\\
AGH University of Science and Technology\\
e-mail: {\tt pszwed@agh.edu.pl}   
}
\date{\today}

\maketitle

\begin{abstract}
In this paper we propose an extension to the Fuzzy Cognitive Maps (FCMs)  that aims at aggregating a number of reasoning tasks into a one parallel run. The described approach consists in replacing real-valued activation levels of concepts (and further influence weights) by random variables. Such extension, followed by the implemented software tool, allows for determining ranges reached by concept activation levels, sensitivity analysis as well as statistical analysis of multiple reasoning results.  
We replace multiplication and addition operators appearing in the FCM state equation by appropriate convolutions applicable for discrete random variables. To make the model computationally feasible, it is further augmented with aggregation operations for discrete random variables. We discuss four implemented aggregators, as well as we report results of preliminary tests.

\end{abstract}

\section{Introduction}
\label{sec:intro}

Fuzzy Cognitive Maps (FCMs) are a well-known tool for modeling and qualitative analysis of various problems \cite{Kosko1992,Aguilar2005,Papageorgiou2011,PapageorgiouSalmeron2014}. They use 
a simple representation of knowledge in a form of a directed graph, in which vertexes are interpreted as concepts and edges attributed with weights as causal relationships. FCMs exhibit certain similarity to neural networks as regards structural properties and reasoning techniques. However, they are considered a semantic modeling tool: concepts, which are typically identified by experts, occur in the problem domain, and weights specifying influences can be explained based on experts knowledge or data used in learning process.  
Below we provide a short theoretical introduction to FCMs. 

Let $C=\{c_1,\dots,c_n\}$ be a set of FCM concepts. A state of the FCM is an $n$-dimensional vector of concept activation levels  ($n = |C|$), which, depending on a setting, are real values from $[0,1]$ or $[-1,1]$.

Causal relations between concepts are represented in an FCM by edges and assigned weights. 
A positive weight of an edge linking two concepts $c_j$ and $c_i$ models a situation, where an increase of the level of $c_j$ results in a growing $c_i$; a negative weight is used to describe the opposite rapport. 
Often, during modeling an ordinal scale of linguistic weights is employed. The symbolic names are then mapped onto a set of real values from the interval $[-1,1]$, e.g. \emph{strong\_negative} ($-1$), \emph{negative} ($-0.66$), \emph{medium\_negative} ($-0.33$), \emph{neutral} ($0$), \emph{medium\_positive} ($0.33$), \emph{positive} ($0.66$), \emph{strong\_positive} ($1.0$).

A representation of FCM that is used during reasoning is an $n\times n$ influence matrix $W = [w_{ij}]$. A value of an element $w_{ij}$ corresponds to a weight of the edge linking concepts $c_j$ and $c_i$ (0 values are used, if there is no link). 
Reasoning with FCM consists in building a sequence of states: 
$\alpha =  A(0), A(1), \dots, A(k), \dots$ 
starting from an initial vector $A(0)$ of concepts activation levels. Successive elements are calculated according to the formula (\ref{eq:state-equation}). In the $k+1$ iteration the vector $A(k)$ is multiplied by the influence matrix $W$, then the resulting activation levels of concepts are mapped onto the assumed range by means of an \emph{activation} (or \emph{splashing}) function $S$.  

\begin{equation}
\label{eq:state-equation}
A_i(k+1) = S(\sum\limits_{j=1}^n w_{ij}\,A_j(k))
\end{equation}


Commonly used activation functions include bivalent or trivalent step functions, a linear function with cutting off values beyond $[-1,1]$, various sigmoidal functions including the logistic function  or the hyperbolic tangent. In our experiments we have also used another S-shaped function $S_{exp}$ defined as  $S_{exp}(x)= 1-\exp(-mx)$ if $x \geq 0$ and $\exp(-mx)-1$, if $x < 0$. The coefficient $m$ allows to adjust the curve slope.

Basically, a sequence of consecutive states $\alpha =  A(0), A(1),\dots, A(k),\dots$ is infinite. However, it was shown that after $k$ iterations, where $k$ is a number close to the rank of matrix $W$, a steady state is reached or a cycle occurs. 
%
%

The sequence of states $\alpha$ can be interpreted in two ways.
Firstly, it can be treated as a representation of a dynamic behavior of the modeled system.  In this case there exist  implicit temporal relations between consecutive system states and the whole sequence describes an evolution of the system in the form of a \emph{scenario}. 
Under the second interpretation, the sequence represents a non-monotonic fuzzy inference process, in which selected elements of a steady state are interpreted as reasoning results. 
In both cases results of reasoning with FCMs can be interpreted only qualitatively, as they strongly depend on granularity of weights and the activation function used.
For example, rather a few scenario steps indicating the predicted development tendencies should be considered or, in the case of reasoning, meaningful results can be related to the ordering of activation levels in a steady state and their proportions.

Both applications of FCMs usually involve executing them for multiple combinations of initial activation levels of concepts: either to test several scenarios starting from various initial states, perform sensitivity analysis or to validate reasoning results for several inputs.   

In this paper we propose an extension to the FCM model, named FCM4DRV, that aims at aggregating a number of reasoning tasks into a one parallel run. The described extension was motivated by the problem of qualitative evaluation of reasoning results for an FCM model of risks related to IT security \cite{SzwedSkrzynskiGrodniewicz2013,SzwedSkrzynski14,SzwedSkrzynskiChmiel14}, however, it is rather a general one, than tailored for a specific purpose. 
The idea behind FCM4DRV consists in replacing real-valued activation levels of concepts (and further influence weights) by random variables. Such extension, followed by the implemented software tool, allows for statistical analysis of multiple reasoning results.  We replace multiplication and addition operators appearing in FCM state equation by appropriate convolutions applicable for discrete random variables. To make the model computationally feasible, we further augment it with aggregation operations for discrete random variables. We discuss four implemented aggregators, as well as we report preliminary test results for an FCM model, which was examined in our previous work \cite{SzwedFCM2013}.                  

The paper is organized as follows: next Section~\ref{sec:related-works} discusses related works and gives a motivation for FCM extension.  It is followed by Section~\ref{sec:fcm-drv}, which introduces FCM4DRV . 
Next Section~\ref{sec:aggregators} presents four implemented aggregators.
Results of experiments are reported in Section~\ref{sec:experiments}. Last Section~\ref{sec:conclusions} provides concluding remarks.

\section{Related works}
\label{sec:related-works}

Fuzzy Cognitive Maps (FCMs) were proposed by Kosko \cite{Kosko1992} as a method for specification and analysis of causal relations between concepts.  
A large number of applications of FCMs were reported, e.g. in project risk modeling \cite{Lazzerini2011},  crisis management and decision making, analysis of development of economic systems and the introduction of new technologies \cite{Jetter2011}, traffic prediction \cite{ChmielSzwed2015}, ecosystem analysis \cite{Ozesmi2004}, signal processing and decision support in medicine. A survey on Fuzzy Cognitive Maps and their applications can be found in  \cite{Aguilar2005} and \cite{Papageorgiou2011}.

Over last 15 years a number of FCM extensions have been proposed. 
Fuzzy Grey Cognitive Maps \cite{Salmeron20123818} use gray numbers (pairs defining interval bounds) as weights in influence matrix. In Intuitionistic FCMs \cite{IakovidisPapageorgiou2011} weigths of influence matrix are also pairs of numbers, the first expresses an impact ($\mu$), the second a hesitation margin ($\pi$). Dynamic Random FCMs \cite{aguilar2004dynamic} introduce probabilities of concept activation, as well as a capability of updating weights during execution.  Other extensions described in \cite{PapageorgiouSalmeron2014} include Rule-based FCMs, Fuzzy Cognitive Networks and Fuzzy Time Cognitive Maps. 
The model of RFCMs (Relational Fuzzy Cognitive Maps) proposed of in \cite{Slon2014} shares to a certain extent features of the discussed FCM4DRV approach. It used fuzzy numbers as concept activation levels and fuzzy relations to define their influences.

In our previous works \cite{SzwedSkrzynskiGrodniewicz2013,SzwedSkrzynski14,SzwedSkrzynskiChmiel14} we have proposed to use FCMs for evaluation of risk related to security of IT systems. FCM models  were hierarchical structures, in which concepts represented assets, risk factors and countermeasures. The FCM reasoning technique was then applied to perform risk aggregation: at first risk factors and countermeasures were combined, then states of low-level assets and their influences allowed to assign utility values to assets placed at higher levels in the hierarchy. 
However, the method faced the problem of correct benchmarking for obtained risk levels, e.g. a question can arise: \emph{how to map a value $0.12$ determined for a certain asset to an ordinal scale of low, medium and high risk}. 
Selection of thresholds supporting such scale can be determined by evaluating numerous combinations of countermeasures. Moreover, preferably it should be based on statistical distribution of system features, e.g. according to best practices some security functions are likely to be implemented more often than others.  
One of the motivating applications of described here extension to the FCM model was to facilitate  thresholds selection, based on percentile ranks of concept activation levels.

\section{Fuzzy cognitive maps for discrete random variables}
\label{sec:fcm-drv}

Random variable $X\colon \Omega\to E$ is a function that maps a sample space $\Omega$ into a measurable space $E$. The sample space represents a set of experiments, measurements or events. A random variable $X$ is called \emph{discrete}, if $E$ is finite or countable, otherwise it is \emph{continuous}. Probability function $p(x) = P(X=x)$, assigns a value from $[0,1]$ to an outcome of a random variable $X$. Moreover, it is required for the sum (or integral) of $p(x)$ over $x\in E$ to be equal $1.0$.      

In the presented model random variables are used as concept activation levels of Fuzzy Cognitive Maps.
Although we assume, that their values lay within a certain interval $[min,max]\subseteq \mathbb{R}$, e.g. $min=-1$, $max=1$, we consider them discrete, i.e. their ranges $E$ are finite. In particular, we represent them as discrete probability mass functions $p \colon E \to [0,1]$, as well as apply addition and multiplication operators appropriate rather for discrete random variables than continuous.
Special cases of random variables are \emph{singletons}, which have a single value $c$: $E_c=\{c\}$ occurring with the probability  $p_c(c)=1$.

\subsection{Arithmetic of discrete random variables}
Let $X$ and $Y$ be two independent discrete random variables (DRV) with probability distributions $p_x(x)$ and $p_y(z)$.
Their sum $Z = X \oplus Y$ is also a random variable with the range  $E_z=\{z\colon \exists (x,y)\in E_x \times E_y\wedge z=x+y \}$ and
whose probability distribution $p_z(z)$ is a \emph{convolution}
of $p_x(x)$ and $p_y(y)$ (\ref{eq:convoloution-sum})
\footnote{Convolution is often defined as $p_z(z)=\sum_x p_x(x)p_y(z-x)$. Formula (\ref{eq:convoloution-sum}) is an equivalent definition. }.

\begin{equation}
p_z(z) = \sum_{x\in E_x}\sum_{\substack{y\in E_y\\z=x+y}} p_x(x) p_y(y) 
\label{eq:convoloution-sum}
\end{equation}

\noindent Similarly, a product $V = X\otimes Y$ is a random variable with the range $E_v=\{v\colon \exists (x,y)\in E_x \times E_y\wedge v=x\cdot y \}$ and a probability distribution given by 
(\ref{eq:convoloution-mul})
\begin{equation}
p_v(v) = \sum_{x\in E_x}\sum_{\substack{y\in E_y\\v=x\cdot y}} p_x(x) p_y(y) 
\label{eq:convoloution-mul}
\end{equation} 
   

Let $S:\mathbb{R}\to \mathbb{R}$ be a scalar function. It induces a function $\hat{S}\colon \{X_i\} \to \{X_i\}$ in the domain of DRVs $\{X_i\}$.
Variable  $Y = \hat{S}(X)$ is defined as:

\begin{equation}
\begin{array}{ccc}
E_y=\bigcup_{x\in E_x}S(x) &\text{ and }& 
p(y)=\sum_{\substack{x\in E_x\\y=S(x)}} p(x)\end{array}\\
\label{eq:activation-function-drv}
\end{equation}



\subsection{Formulation of FCM4DRV}

In FCM4DRV, which is an extension to the basic FCM model, concept activation levels are represented by discrete random variables (DRVs), similarly the influence matrix $W$ is an $n \times n$ matrix of DRVs and FCM states are $n$-dimensional vectors of DRVs. Let us observe, that a classical FCMs can be considered a special case of the extended model, where all DRVs are just signletons (single values with assigned probability $1$). Under such assumptions, the FCM state equation (\ref{eq:state-equation}) can be rewritten as in (\ref{eq:fcmdrv-state}) using defined earlier summation $\oplus$ and multiplication $\otimes$ operators, as well as an activation function $\hat{S}$ defined in the domain of DRVs.     

\begin{equation}
A_i(k+1) = \hat{S}	(w_{i1}\otimes A_1(k)\oplus w_{i2}\otimes A_2(k)\oplus \dots \oplus w_{in}\otimes A_n(k))
\label{eq:fcmdrv-state}
\end{equation}

Analogously to the classical model, execution of FCM4DRV produces a sequence of states $\alpha =  A(0), A(1), \dots, A(k), \dots$, whose convergence can be checked based on selected distance measure for DRVs.  

Unfortunately, in most practical situations calculation of a new FCM state with formula (\ref{eq:fcmdrv-state}) is computationally unfeasible. Consider a simple case of $n\times n$ influence matrix $F$ of singletons (i.e. a real-valued matrix) and an initial state vector $A_0$ of DRVs, each having ranges of $k$ elements. Then, in the worst case the DRV ranges in $A_1$ will comprise $k^n$ elements, $k^{2n}$ elements for $A_2$, $k^{in}$ for $A_i$  and so on.
If we assume quite a reasonable values $k=100$ and $n=10$, then probably the calculation of $A_1$ with $10\cdot 100^{10}=10^{21}$ mapping elements would fail.        

To handle this problem we introduce additional \emph{aggregation} operation  into the state equation that 
is applied to partial results obtained during evaluation of expression appearing on the right side of the state equation (\ref{eq:fcmdrv-state}). An aggregation function $\hat{G}$ converts an input DRM $X$ into a smaller (i.e. having less numerous mapping) variable $Y = \hat{G}(X)$. It is expected that the number of elements appearing in the range of $Y$  is bounded by a selected positive integer: $|E_Y|<k$ and  certain equivalence criteria are satisfied $Y\approx X$. 

For the discussed above example, where $n=10$ and $k=100$ the state representation will comprise at most $n\cdot k=1000$ elements and the number of operations required to compute the next state will be bounded by $n \cdot (n-1)\cdot k^2 = 9\cdot 10^6 $ ($n$ rows, $n-1$ occurences of $\oplus$ operator, $k^2$ -- complexity of convolution).       

The FCM state equation extended by aggregation function $\hat{G}$ is given by (\ref{eq:fcmdrv-state-aggr}). It reflexes the most natural order of evaluating expressions (from left to right).

\begin{equation}
\begin{array}{lcl}
A_{i1}(k+1)&=&f_{i1}\otimes A_1(k)\\
A_{i2}(k+1)&=&\hat{G}(A_{11}(k) \oplus (f_{i2}\otimes A_2(k)))\\
\vdots\\
A_{in}(k+1)&=&\hat{G}(A_{1n-1}(k) \oplus (f_{in}\otimes A_n(k)))\\
\text{and finally:}\\
A_i(k+1) = \hat{S}(A_{in}(k+1))
\end{array}
\label{eq:fcmdrv-state-aggr}
\end{equation}

Equivalence relation $Y\approx X$ for random variables can be based on various measures, e.g. equality or expected values $E(X) = E(Y)$ or distances between two distributions, like 
earth mover's distance  \cite{rubner2000earth} or Kolmogorov-Smirnov distance \cite{2006practical}. At this point, however, we did not make attempt do qualitatively evaluate aggregation functions and analyze their influence on  states reached during reasoning. Instead in the next Section~\ref{sec:aggregators}, we describe a few prototype aggregation methods that were developed and used during experiments.     
    
%

\section{Aggregators}
\label{sec:aggregators}
For a discrete random variable $X$ its probability mass function $p_X\colon E_X \to [0,1]$ is actually represented as a set of pairs: $p_X = \{(x,p)\colon x \in  E_X \wedge p\in [0,1]\}$. In our experiments $E_X$ was finite and hence bounded: $E_X \subset [x_{min},x_{max}]$.
The basic idea behind at least three aggregators described in this section consists in performing one-dimensional clustering. Values $x\in E_X$ laying close are grouped into clusters $E_{X1},\dots, E_{Xi},\dots E_{Xk}$ and each cluster $E_{Xi}$ is replaced by a single pair $(v_i,p_i)$. The method of establishing $v_i$ and $p_i$ depends on algorithm, typically $v_i$ is obtained by kind of averaging values in $E_{Xi}$ and $p_i$ by summing probabilities.

\subsection{Simple k-means}
\label{subsec:k-means}

Simple k-means is an adaptation of well-known k-means clustering algorithm \cite{witten2005data}. The main difference is that initial centroids are not randomly selected, but evenly distributed within the range $[x_{min},x_{max}]$. For the resulting $p_V=\{(v_i,p_i)\}$ elements  $v_i$ are cluster centers and $p_i$ is a sum of probabilities assigned to elements $E_{Xi}=\{x_{ij}\}$ forming a cluster: 
$p_i=\sum_{x\in E_{Xi}}p(x)$. The method does not assure that the mean value of $X$ will be kept by $V$. In spite of this, during experiments $E(X)$ and $E(V)$ occurred to be quite close.  



\subsection{DBSCAN}
\label{subsec:dbscan}

DBSCAN (Density-based spatial clustering of applications with noise) is a widely used clustering algorithm \cite{witten2005data} characterized by low complexity $O(n\,log n)$. It is controlled by two parameters $\epsilon$ -- minimal distance between data points forming a neighborhood and $\lambda$ -- minimal cluster size. During algorithm execution points, whose neighborhood size is smaller than  $\lambda$ are rejected as outliers. On the other side, neighborhoods having at least $\lambda$ elements are converted to clusters and further expanded. Outcome of the algorithm, including the number of clusters, depends on established values of $\varepsilon$ and $\lambda$. 

The discussed aggregator has been based on DBSCAN implementation in JavaML  library \cite{abeel2009java} with the following parameters:  $\varepsilon = \frac{x_{max}-x_{min}}{k}$, where $k$ is an upper limit on the number of resulting clusters  and $\lambda = 6$. After running the clustering algorithm, 
values belonging to the clusters  
$E_{Xi}$ were converted to pairs $(v_i,p_i)$ according to formula (\ref{eq:dbscan}). 


\begin{equation}
\begin{array}{lcl}
p_i = \sum_{x\in E_{Xi}}p_X(x)&
\text{and}&
v_i=\frac{1}{p_i}\sum_{x \in E_{Xi}}x \cdot p_X(x)\\
\end{array}
\label{eq:dbscan}
\end{equation}

\subsection{UniBins aggregator}
\label{subsec:unibins}

\begin{figure}[!ht]
\centering
\includegraphics[width=0.60\columnwidth]{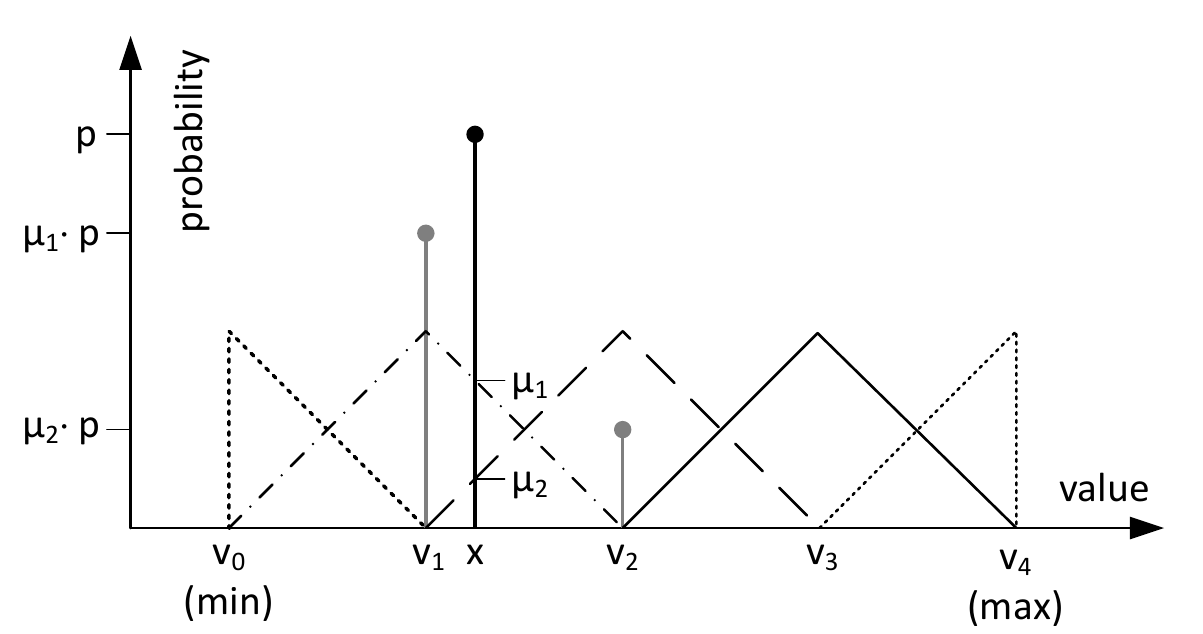}
\caption{UniBins aggregator. Value $x$ with the probability $p$ laying between $v_1$ and $v_2$ contributes $\mu_1\cdot p$ to probability of $v_1$ and $\mu_2\cdot p$ to probability of $v_2$. Factors $\mu_1$ and $\mu_2$ are determined according to triangle membership functions around $v_1$ and $v_2$ marked with dash-dot and dashed lines.}
\label{fig:fcmdrv-unibins}
\end{figure}

UniBins agregator divides the range $[x_{min},x_{max}]$ of an input variable $X$ into $k$ uniformely distributed bins represented by values $v_{0},\dots,v_{n-1}$. Bins borders are fuzzy and 
a level, at which an input element $x$ can be assigned to a bin is quantitatively described by a bin's membership function. This concept is illustrated in Fig.~\ref{fig:fcmdrv-unibins}.

\subsection{Percentile rank aggregator}
\label{subsec:percentile}

\begin{figure}[!h]
\centering
\begin{tabular}{cp{0.1cm}c}
\includegraphics[scale=0.70]{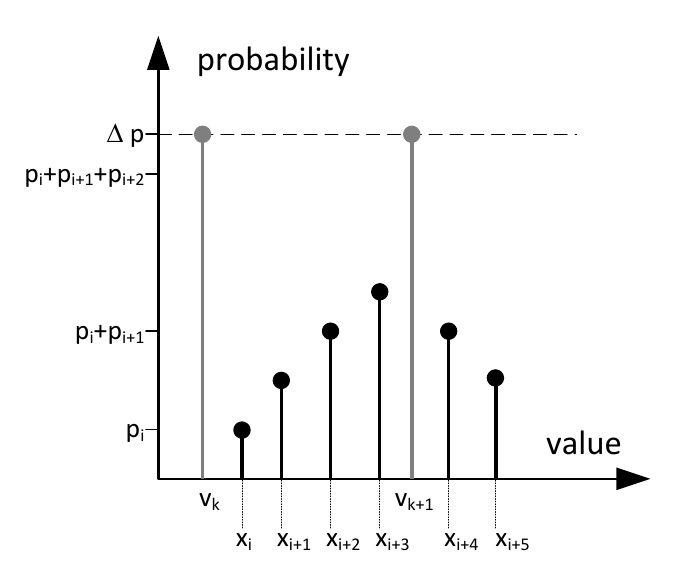} &  & \includegraphics[scale=0.70]{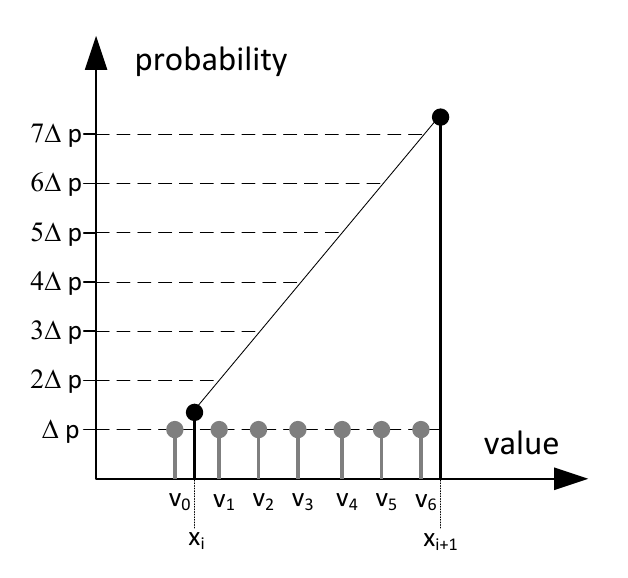} \\ 
(a)  &  & (b)  \\ 
\end{tabular} 
\caption{PercentileRank aggregator:  (a) multiple input values aggregated into one (b) significant change of amplitude resulting in multiple output elements}
\label{fig:fcmdrv-pctrank}
\end{figure}

Percentile rank aggregator assigns equal probability $\Delta p=1/k$ to each output value, while preserving the percentile ranks of input distribution. The assumed granulation level is $\Delta p$. The basic algorithm idea is shown in Fig.~\ref{fig:fcmdrv-pctrank}a. Let us analyze  the sequence $(x_i,p_i),\dots,(x_{i+3},p_{i+3})$. As $p_i+p_{i+1}+p_{i+2} < \Delta p < p_i+p_{i+1}+p_{i+2}+p_{i+3}$ the value $v_{k+1}$ will be placed between $x_{i+3}$ and $x_{i+4}$. The exact position depends on $\Delta p -p_i-p_{i+1}-p_{i+2}$, the smaller the value is, the distance between $x_{i+3}$ and $v_{k+1}$ is smaller.

Another feature of the percentile rank aggregator is its capability to produce multiple output values in case of rapid changes of input PMF. This is illustrated in Fig.~\ref{fig:fcmdrv-pctrank}b: placement of output values $v_1,\dots,v_6$ correspond to points of intersection of line linking $x_i$ and $x_{i+1}$ with successive percentile ranks: $2\Delta p, 3 \Delta p, \dots, 7 \Delta p$.

\section{Experiments and results}
\label{sec:experiments}

In this section we present results of experiments conducted with a prototype software tool supporting  FCM4DRV. The software written in Java implements operations on DRVs, defines a number of activation functions and aggregators and conducts FCM reasoning. 

\begin{figure}[!hb]
\centering
\includegraphics[width=0.65\columnwidth]{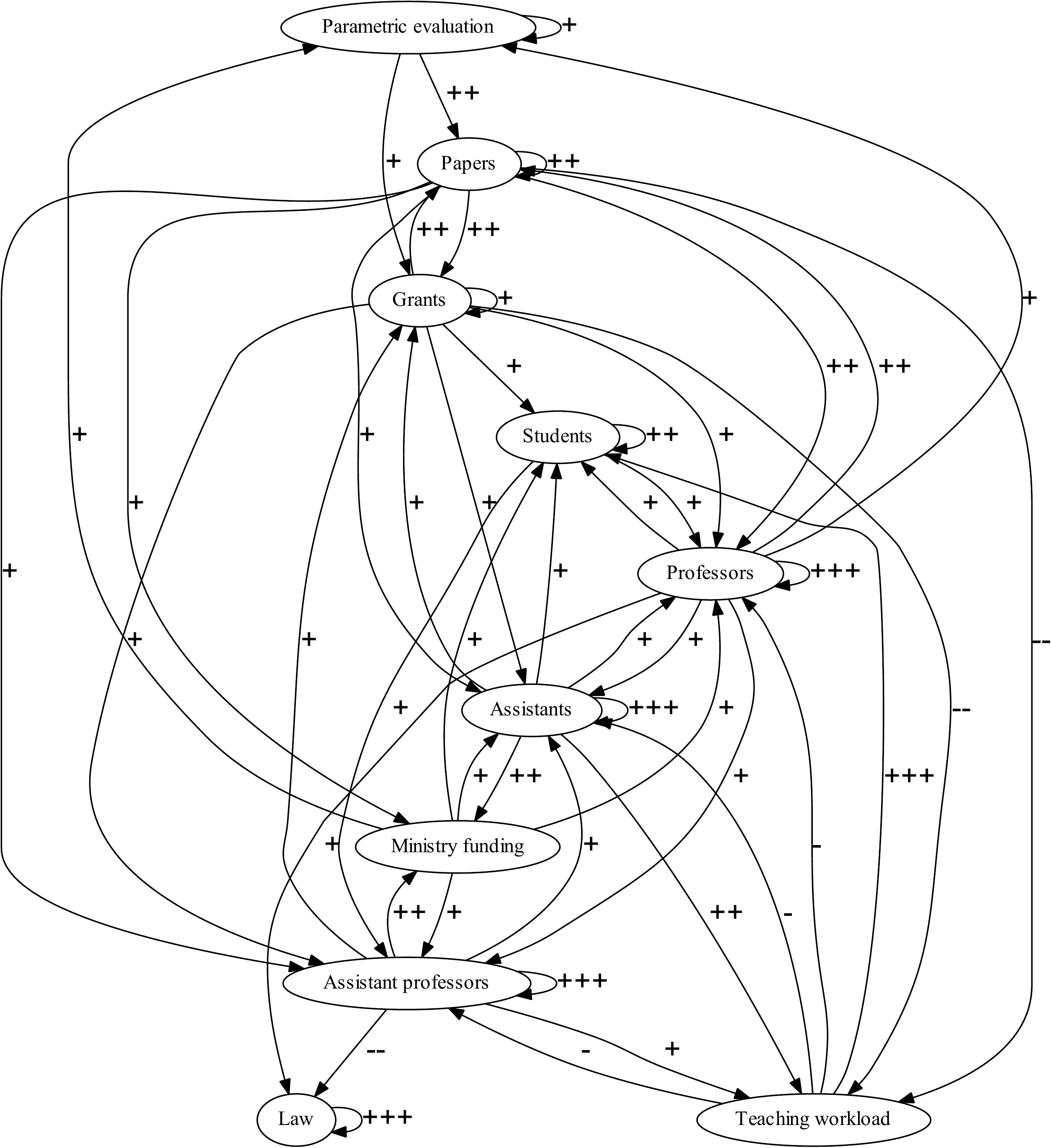}
\caption{Fuzzy cognitive map for analysis of academic units development \cite{SzwedFCM2013}. Linguistic values ($---$, $--$, $-$, $+$, $++$, $+++$) are mapped to numeric  weights $(-1,-0.66,-0.33,0.33,0.66,1)$ .  
}
\label{fig:fcm-model}
\end{figure}

Described further experiments were performed on an FCM model 
that was previously discussed in \cite{SzwedFCM2013}. The map presented in Fig.~\ref{fig:fcm-model} specifies concepts and their influences intended to characterize the domain of academic units, e.g. university departments. Although the model accuracy may be  disputable, it was selected because it was previously quite extensively tested. Moreover, it has easy to perceive semantic, what facilitates the analysis.

The influence matrix used during the experiments  comprised single real values, i.e. singletons with assigned probability $1$. However, all elements of the initial  state vector,  were random variables of 100 values uniformly distributed in the interval $[-1,1]$. The only exception was the input concept 
\emph{Law}, which in each iteration was reset to the single value $1$ with probability $1.0$. The aggregators were configured to keep sizes of DRVs limited to $k=100$.     

All experiments were conducted using Java 8, run on Intel Core i7-2675QM laptop at 2.20 GHz, 8GB memory under Windows 7. The number of iterations was limited to 25, as regardless of activation function and aggregator used all calculations converged to steady states within that bound. Execution times (25 iterations) depended on aggregators: for \emph{Simple k-means} execution times ranged at 9 min 41 seconds, for \emph{DBSCAN} about 6 minutes 51 seconds, for \emph{UniBins} about 5.5 seconds and, finally, 4 seconds in the case of \emph{PercentileRank}.

\begin{figure}[!ht]
\centering
\begin{tabular}{cp{0.1cm}c}
\includegraphics[width=0.4\columnwidth]{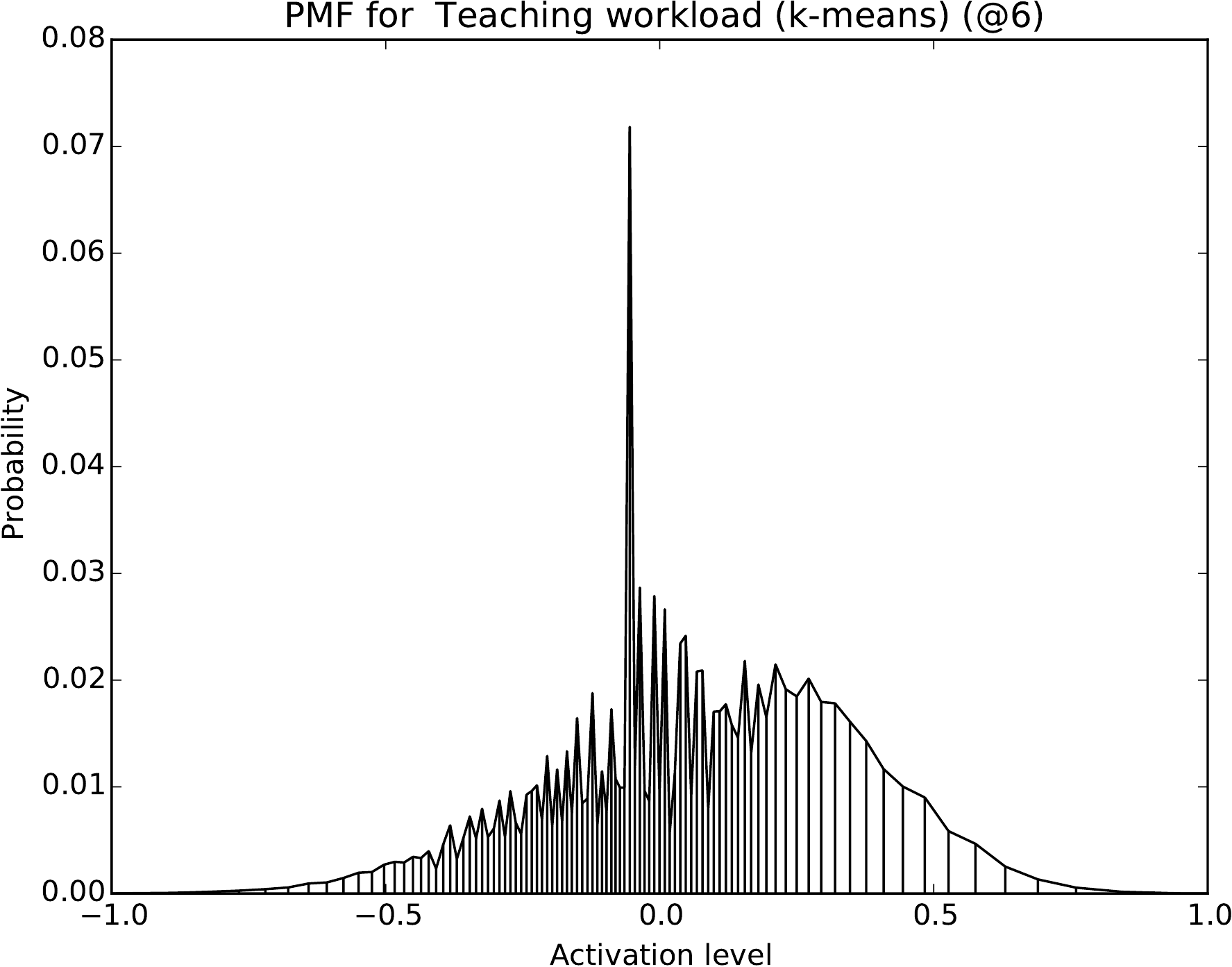}&  &  \includegraphics[width=0.4\columnwidth]{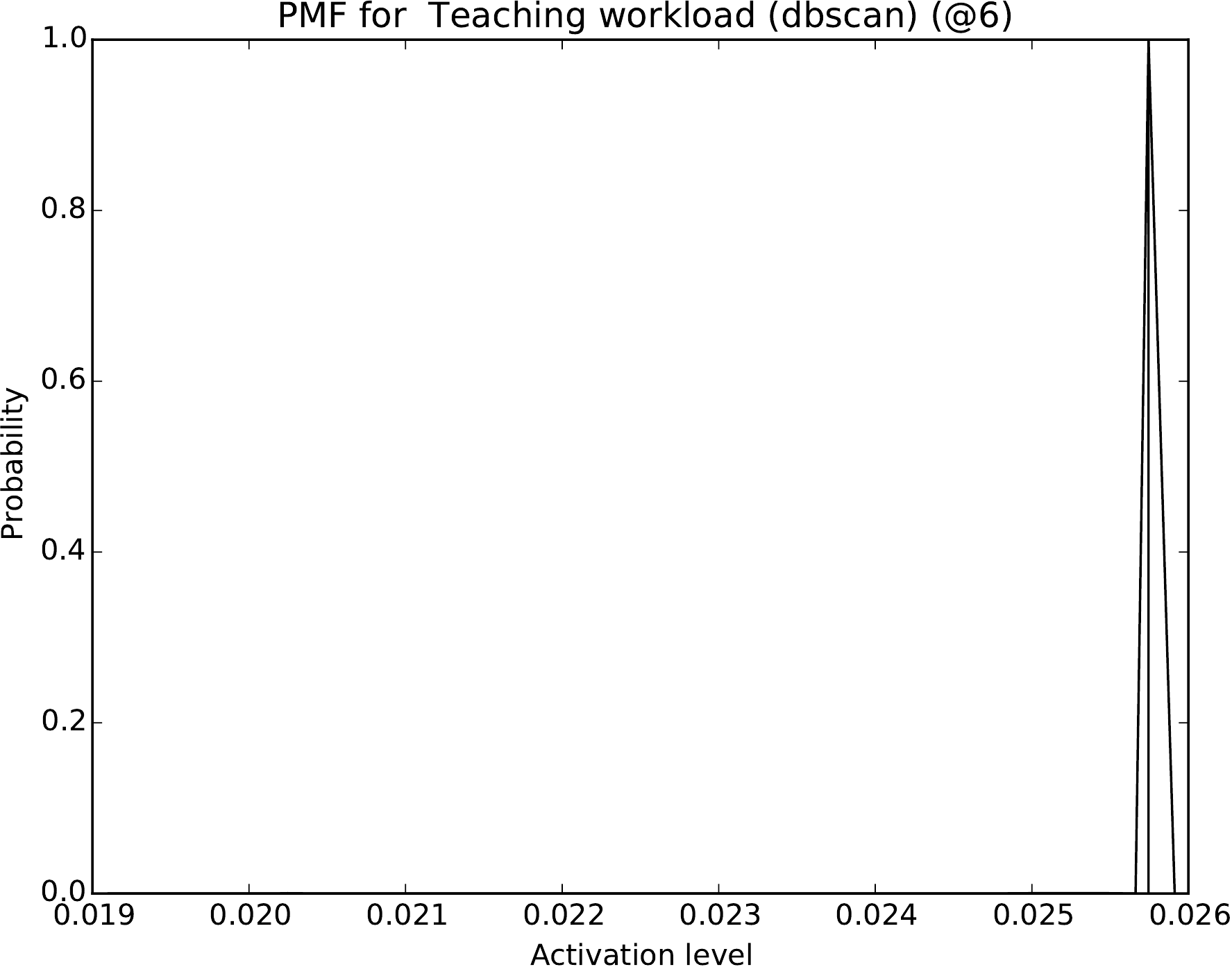}\\ 
(a)  &  & (b)  \\ 
\includegraphics[width=0.4\columnwidth]{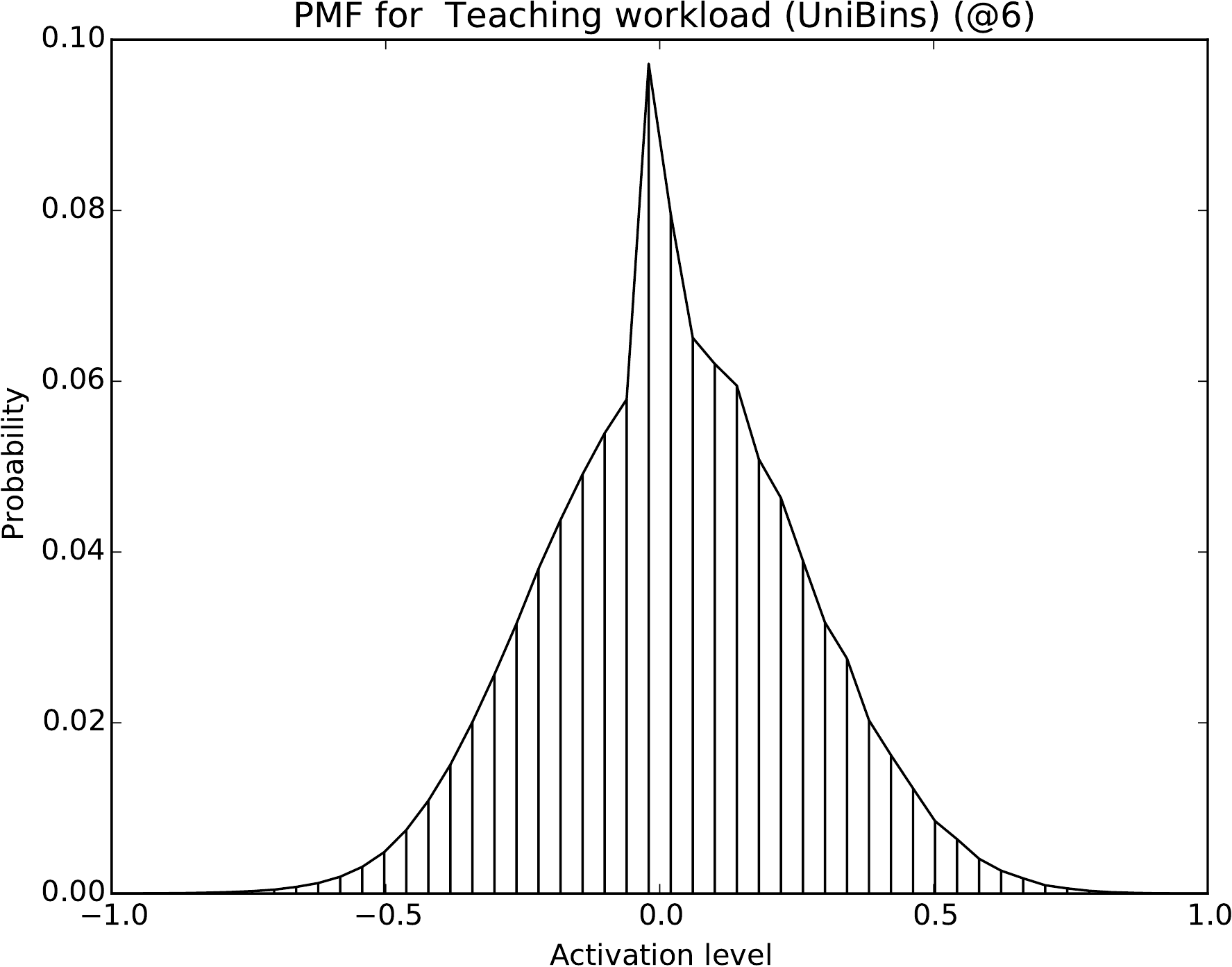}  &  & \includegraphics[width=0.4\columnwidth]{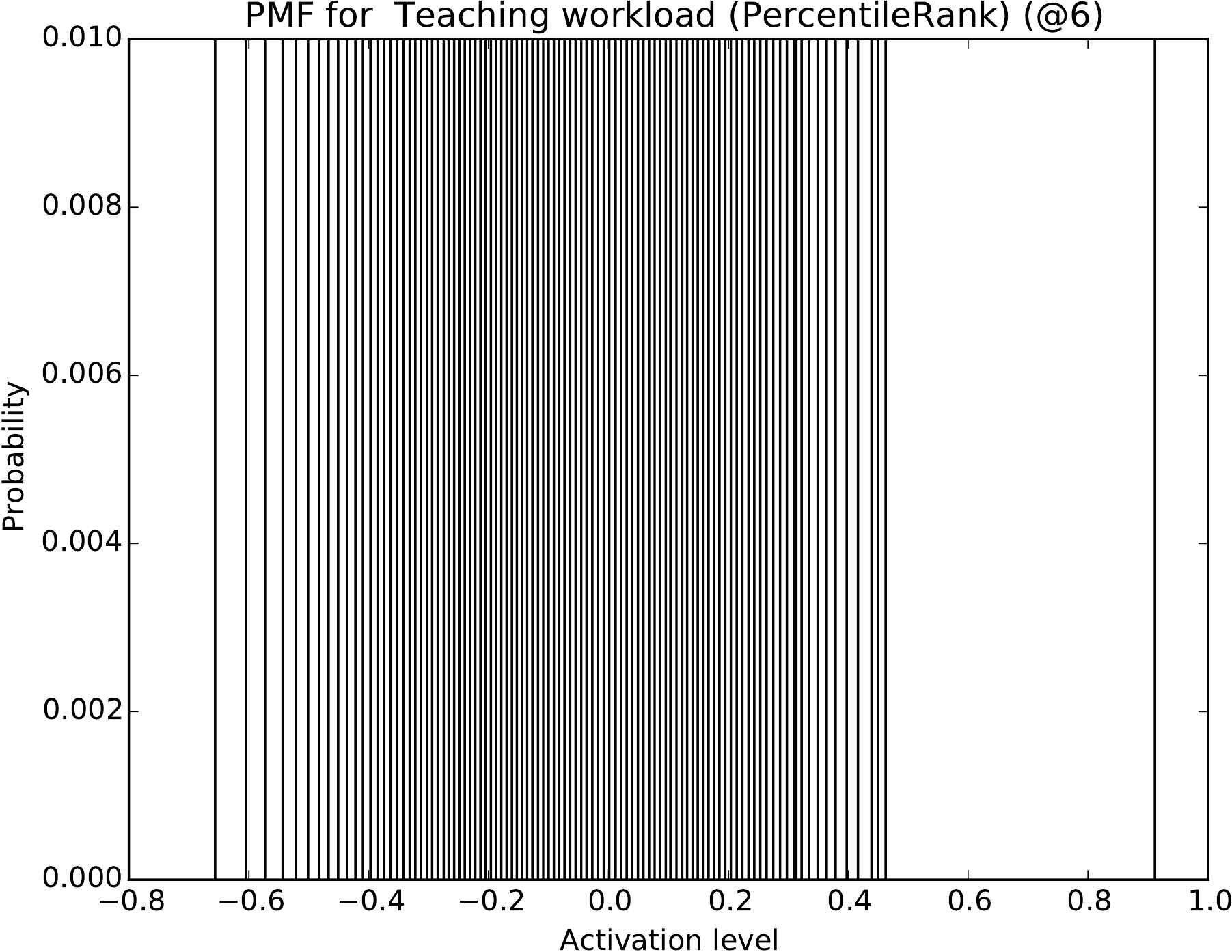} \\ 
(c)  &  & (d)  \\ 
\end{tabular} 
\caption{Comparison of four aggregators: (a) simple k-means (b) DBSCAN (c) UniBins (d) PercentileRank}
\label{fig:experiments-aggregators}
\end{figure}

Fig.~\ref{fig:experiments-aggregators} shows typical probability distributions obtained by applying previously discussed aggregators. We have selected for comparison the concept 
\emph{Teaching workload} at iteration 6. 
Plots (a) and (c) show that observed PMFs are mixtures of 3 (simple k-means) or 2 (UniBins) Gaussian distributions. A typical feature of DBSCAN aggregator is a small number of resulting clusters and in consequence a significant reduction of the number of values occurring in a resulting discrete random variable. In this case the input variable comprising 600 elements was converted to 4 clusters.
The plot (d) shows results of applying PercentileRank aggregator. High amplitudes in other diagrams, e.g. (a) correspond to high frequencies  of values.

\begin{figure}[!h]
\centering
\begin{tabular}{cp{0.1cm}c}
\includegraphics[width=0.4\columnwidth]{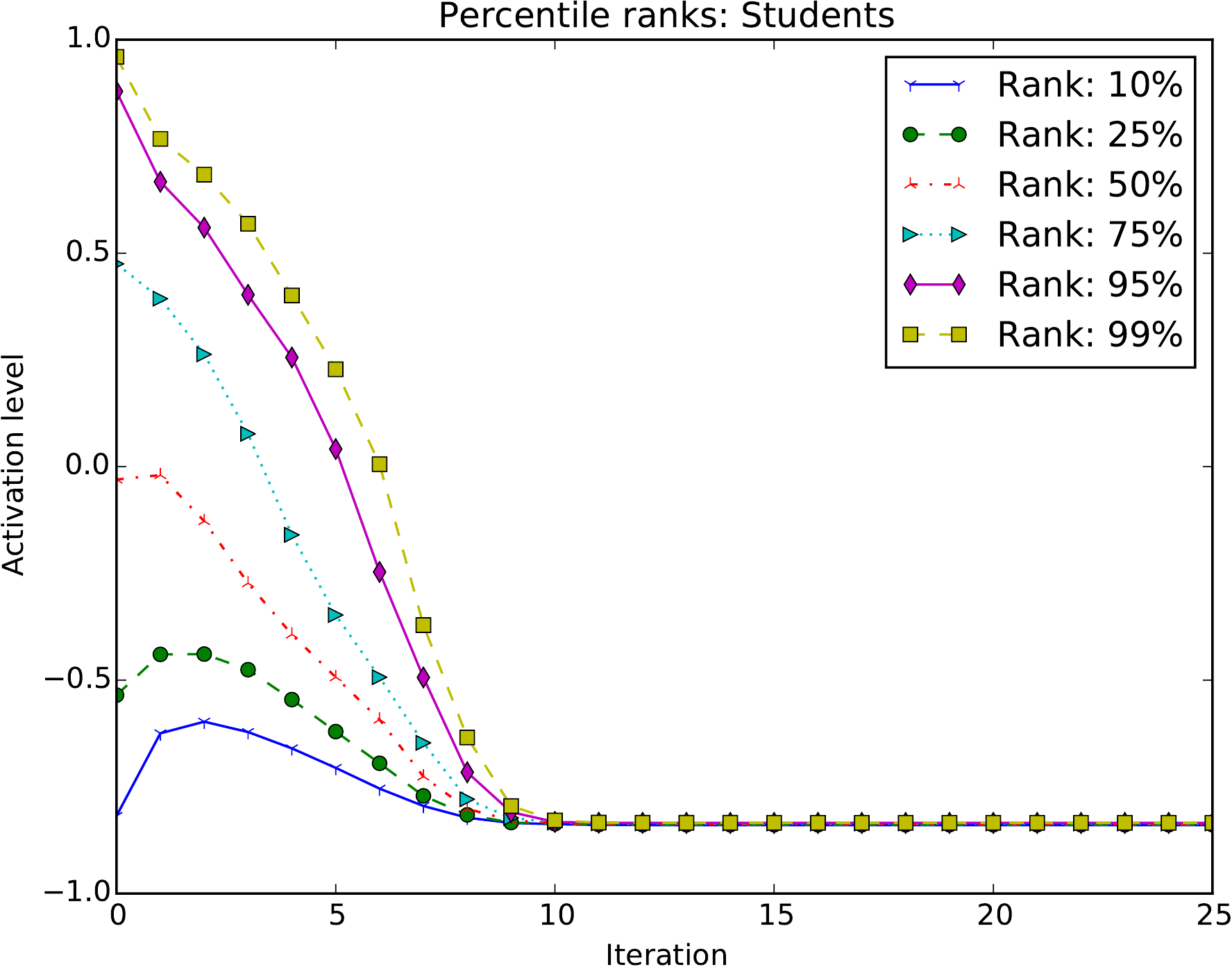} &  & \includegraphics[width=0.4\columnwidth]{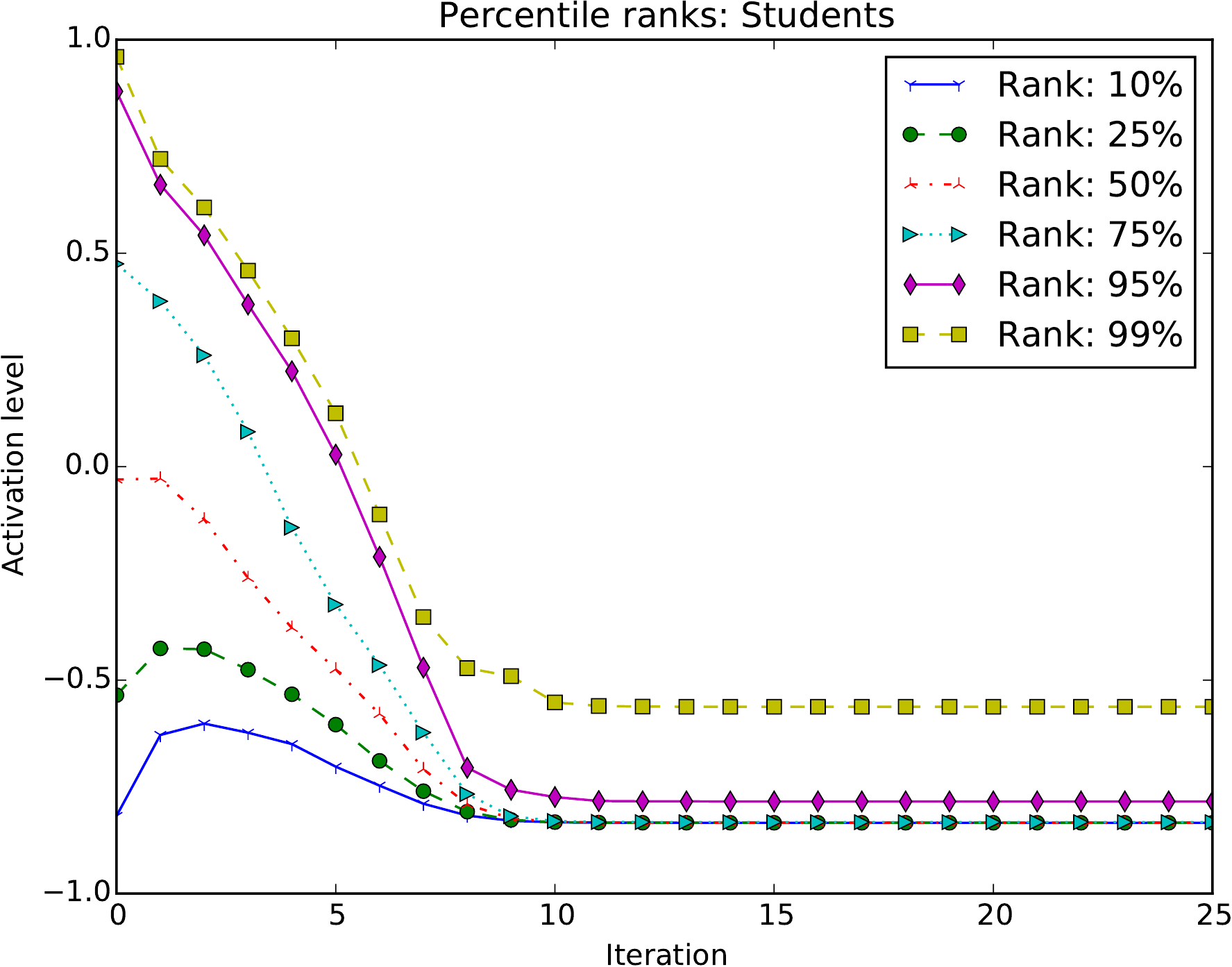} \\ 
\includegraphics[width=0.4\columnwidth]{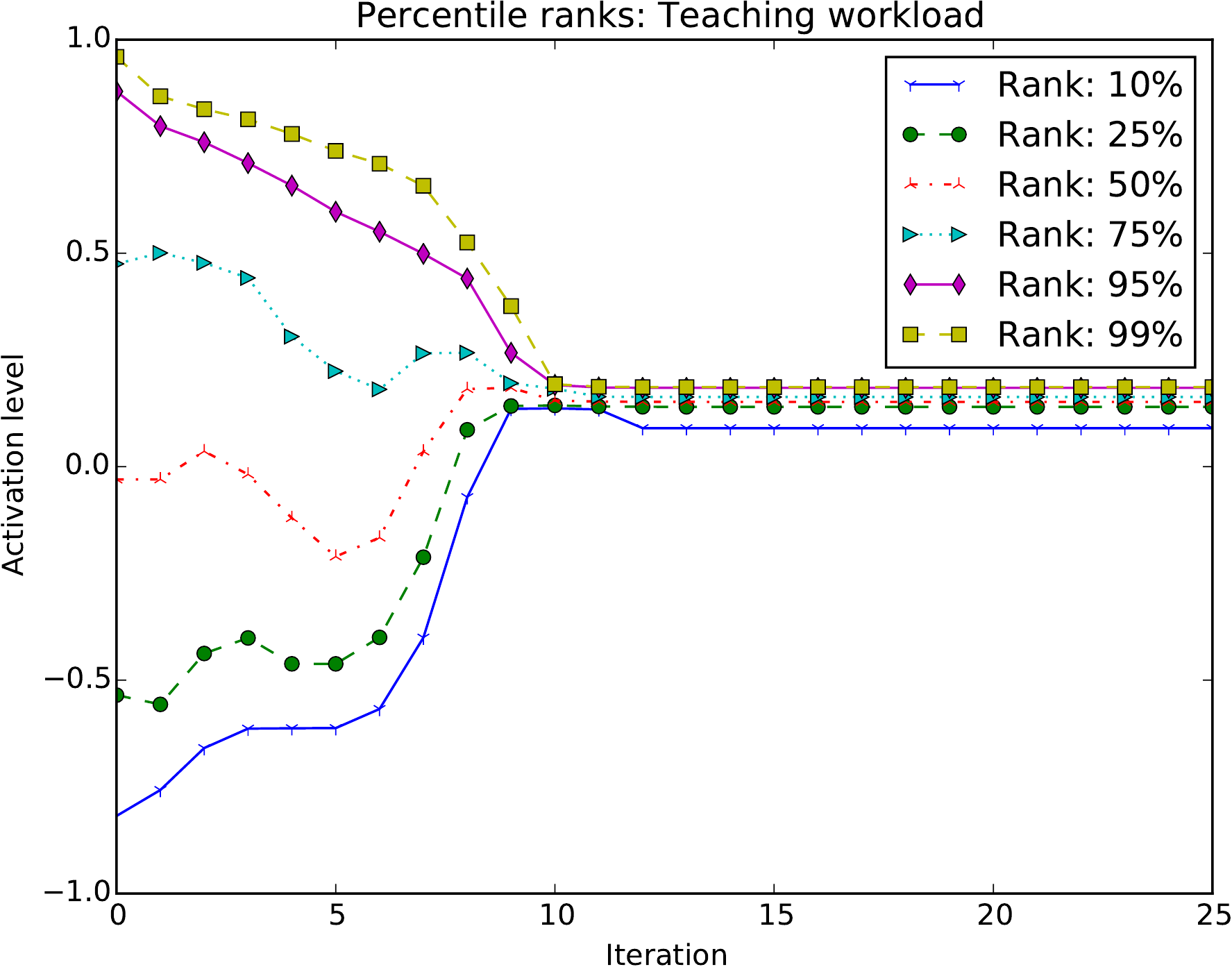} &  & \includegraphics[width=0.4\columnwidth]{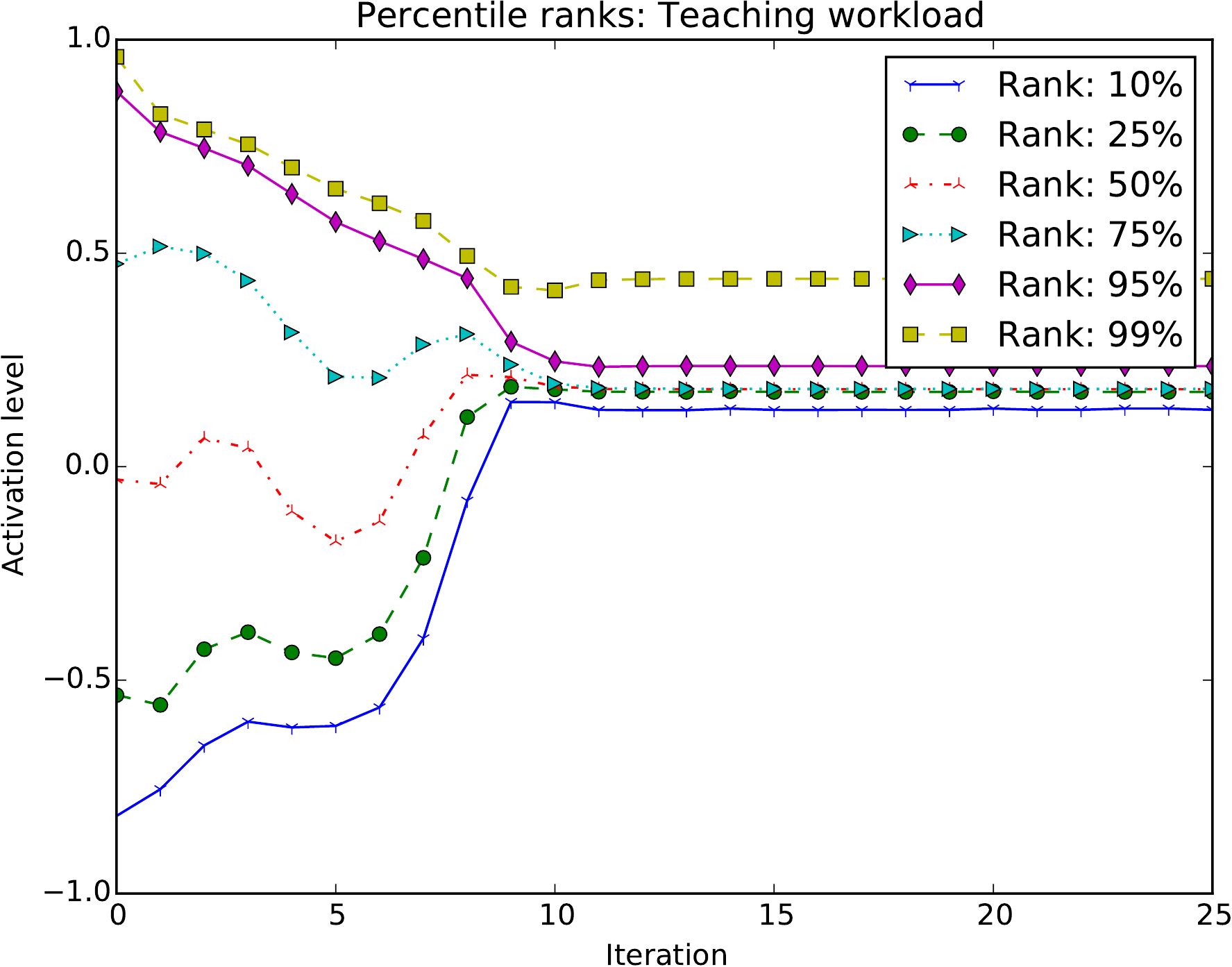} \\ 
\includegraphics[width=0.4\columnwidth]{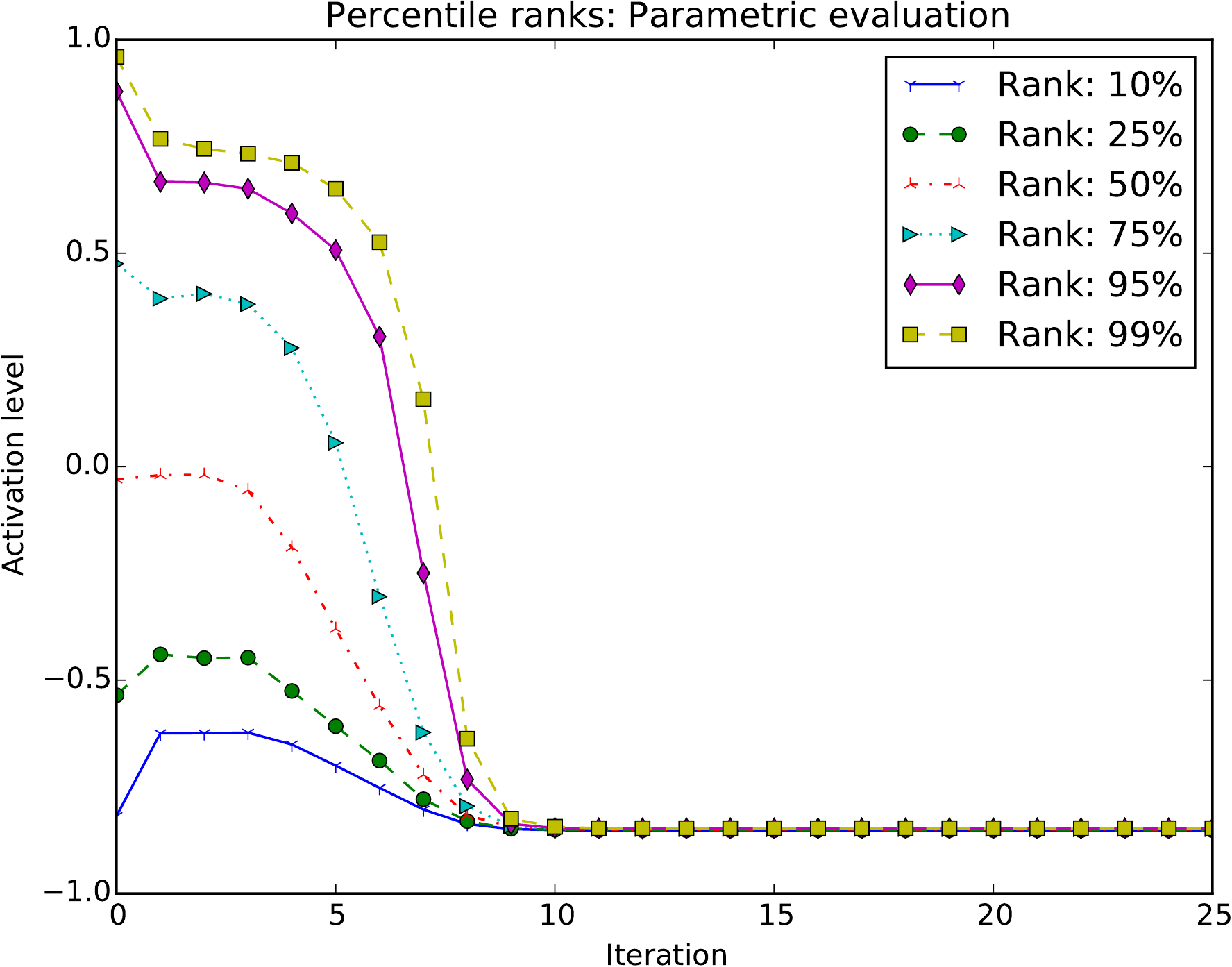} &  & \includegraphics[width=0.4\columnwidth]{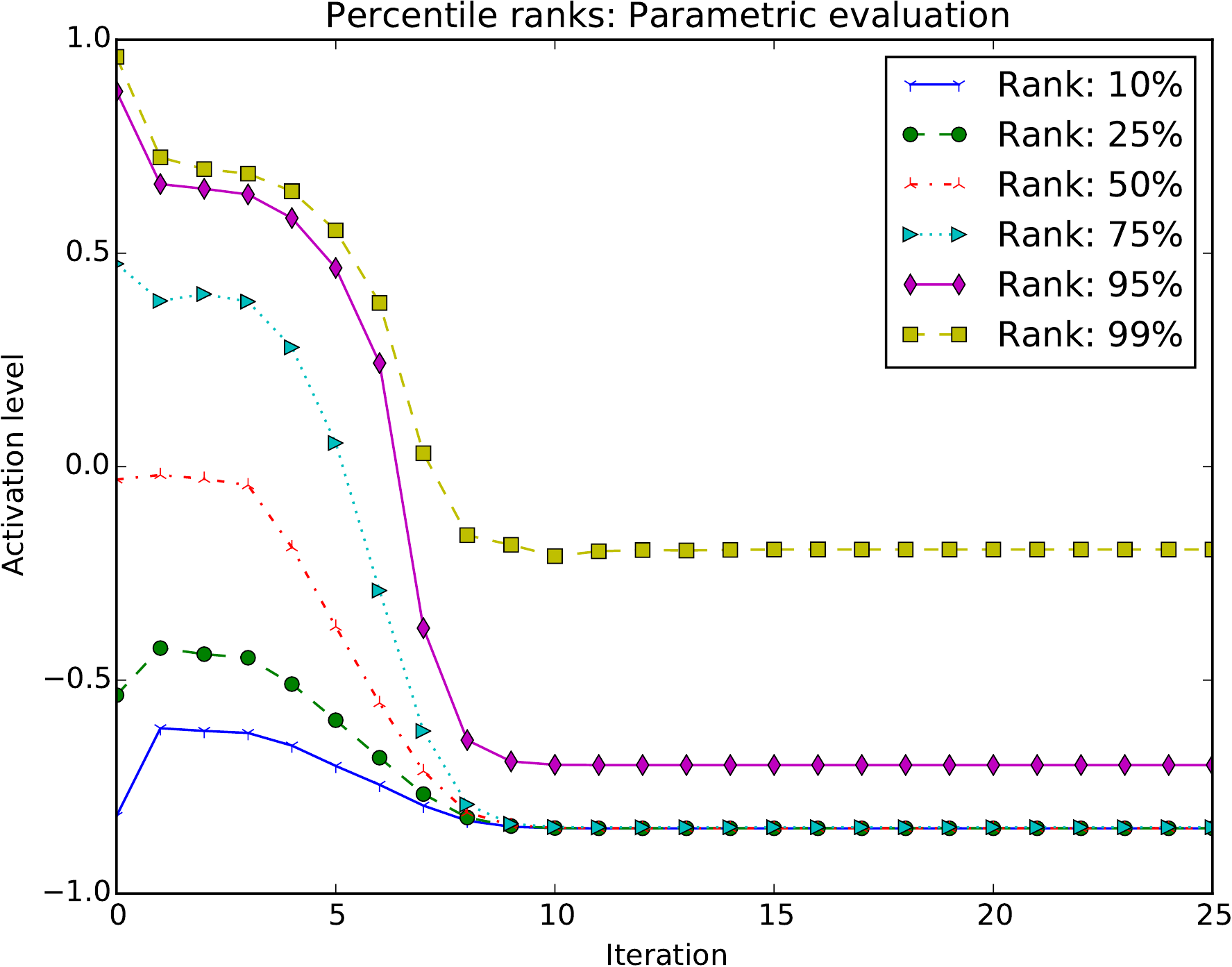} \\ 
\includegraphics[width=0.4\columnwidth]{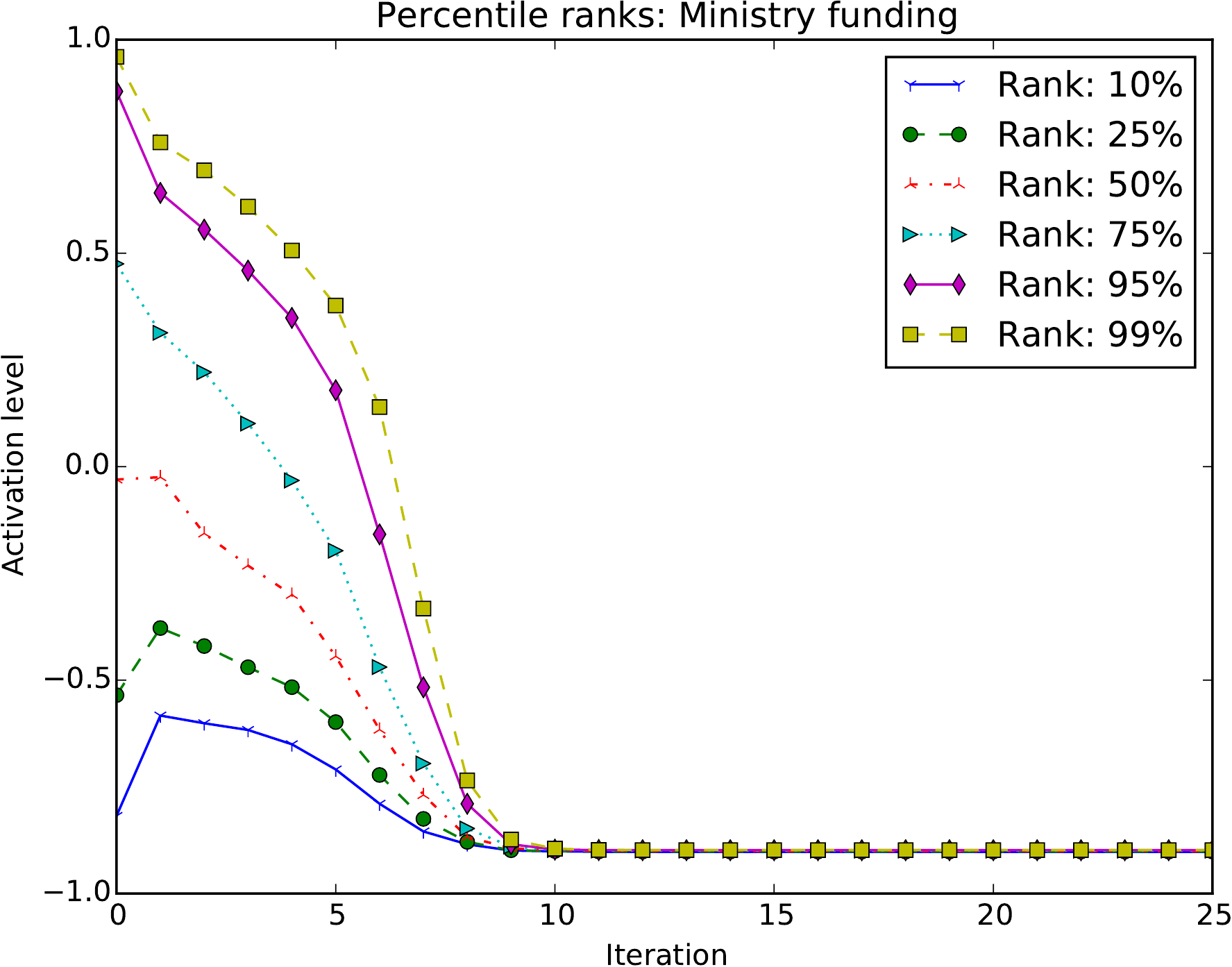} &  & \includegraphics[width=0.4\columnwidth]{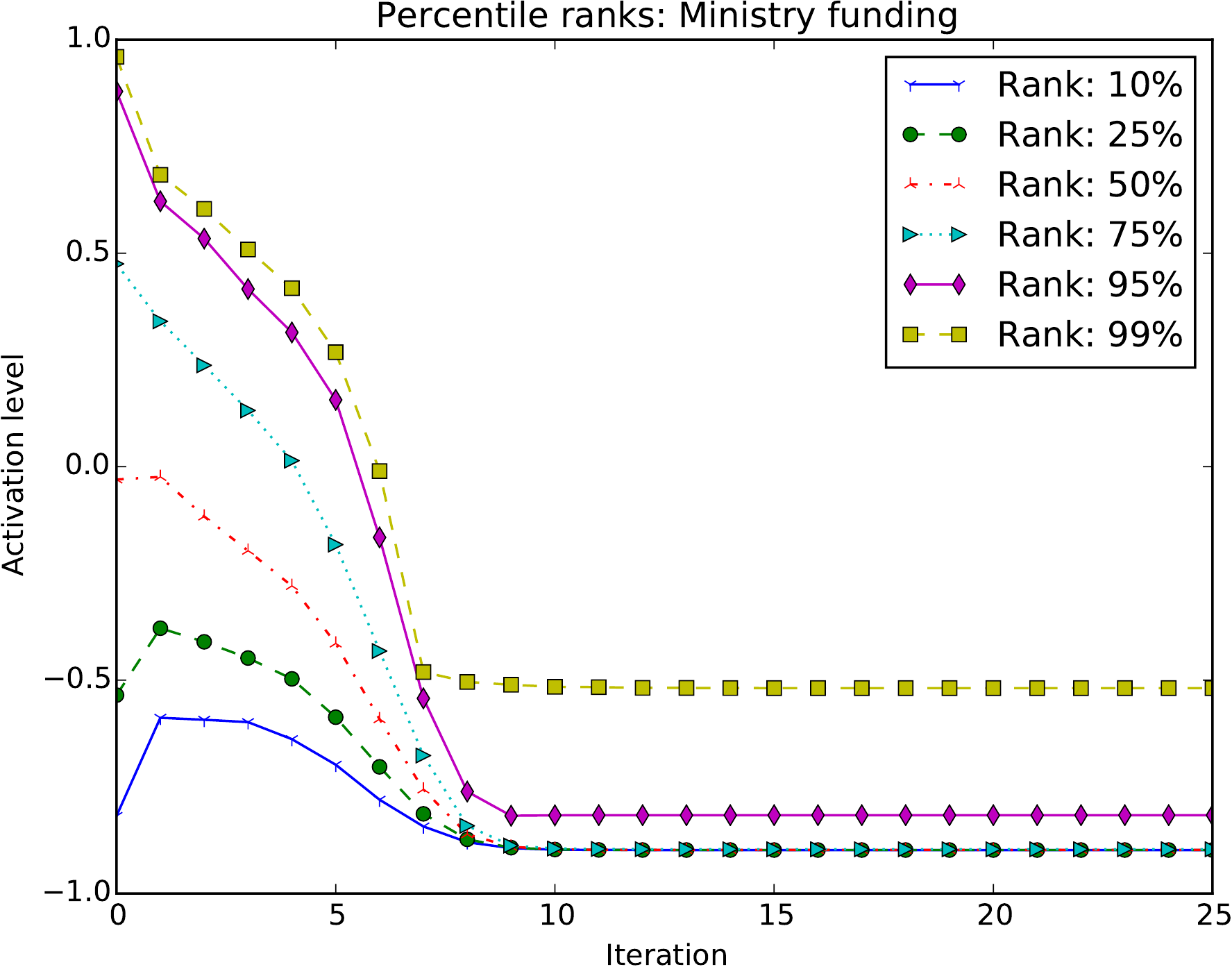} \\ 
(a)  &  & (b)  \\ 
\end{tabular} 
\caption{Comparison of percentile scores obtained (a) UniBins aggregator (b) Percentile Rank aggregator. In both cases exp activation function was used.}
\label{fig:percentile-iteration}
\end{figure}

The primary goal of FCM4DRV is to provide data enabling statistical analysis of ranges reached by concept activation levels during reasoning. Fig.~\ref{fig:percentile-iteration} illustrates such kind of analyzes. It shows how percentile  scores for selected concepts changed over iterations. The left column (a) gives results for UniBins aggregator, while (b) for PercentileRank. In both cases $S_{exp}$ activation function was used. Although the results are qualitatively similar, the plots suggest that the second aggregator is probably more appropriate for analyses related to percentile ranks.

It should be noted that reasoning with FCM4DRV allows only to establish ranges of activation levels, full information on FCM states that can be reached in a classical reasoning process is not available.   
However, as it was mentioned in Section~\ref{sec:related-works}, such outcomes fits our needs  related to benchmarking of risk levels during risk assessment. (In this case FCMs were used for hierarchical aggregation and we were interested in values obtained after $m$ iterations, where $m$ is the hierarchy depth.) 
On the other hand, activation levels reached in a steady state can be interpreted as expected values for a certain initial distribution. In particular reasoning with FCM4DRV can be used for sensitivity analysis focused on a certain concept, e.g. consider an experiment, in which initial values for one concept are uniformly distributed and all other are fixed as singletons.  We may also put forward a claim that theoretically, for experiments similar to the one discussed, results obtained \emph{in the first iteration} may provide enough information to describe predicted tendencies: as initial activation levels of concepts cover their ranges, sets of values determined in the first iteration comprise all possible reasoning outcomes. However, the use of aggregators introduces errors, which were not at this point analyzed.


\section{Conclusions}
\label{sec:conclusions}

In this paper we discuss FCM4DRV, an extension to classical FCM model consisting in replacing concept activation levels with discrete random variables. The proposed model aims at establishing ranges of activation levels reached during  reasoning with FCMs. We were motivated by a particular problem of selecting accurate thresholds during IT security risk analysis with FCM \cite{SzwedSkrzynskiGrodniewicz2013,SzwedSkrzynski14,SzwedSkrzynskiChmiel14}, however, the presented here solution is more general and can be applied for a variety of problems.
The FCM4DRV extension includes augmenting classical FCM state equation with appropriate operators applicable to DRVs, as well as introducing aggregators, special functions that transform DRVs into similar ones, yet less memory consuming and requiring smaller computational effort. 
We implemented a prototype software tool supporting FCM4DRV model and we give results of experiments demonstrating its computational feasibility and typical results.       

We plan to develop features that are still missing: first of all provide tools for assessing similarity measures between DRVs, errors introduced by aggregators, as well as provide analysis on their influence on reasoning results.


%
%
\bibliography{pszwed,FCM,fuzzy,petri,statistics,ml}

\end{document}